\documentclass[10pt]{article}
\usepackage[latin1]{inputenc}
\usepackage[numbers]{natbib}
\usepackage{times,amsfonts,enumerate,amssymb,amsmath,epsfig,subfigure,bm,cite}

\begin{document}
\markboth{Backes \& Bruno}
{Fractal and Multi-Scale Fractal Dimension analysis: a comparative study of Bouligand-Minkowski method}

\title{Fractal and Multi-Scale Fractal Dimension analysis: a comparative study of Bouligand-Minkowski method}

\author{
\bf Andr\'e Ricardo Backes\\
\normalsize Instituto de Ci\^encias Matem\'aticas e de Computa\c{c}\~ao - ICMC,\\ Universidade de S\~ao Paulo - USP, S\~ao Carlos, Brazil\\
backes@icmc.usp.br \\ \\
\bf Odemir Martinez Bruno \\
\normalsize Instituto de Ci\^encias Matem\'aticas e de Computa\c{c}\~ao - ICMC,\\ Universidade de S\~ao Paulo - USP, S\~ao Carlos, Brazil\\
bruno@icmc.usp.br
}
\date{}

\maketitle


\noindent{}{\bf\large Abstract --} 
Shape is one of the most important visual attributes to characterize objects, playing a important role in pattern recognition. There are various approaches to extract relevant information of a shape. An approach widely used in shape analysis is the complexity, and Fractal Dimension and Multi-Scale Fractal Dimension are both well-known methodologies to estimate it. This papers presents a comparative study between Fractal Dimension and Multi-Scale Fractal Dimension in a shape analysis context. Through experimental comparison using a shape database previously classified, both methods are compared. Different parameters configuration of each method are considered and a discussion about the results of each method is also presented.
\\
\noindent{}{\bf\large Keywords --}
Complexity; Shape Analysis; Fractal Dimension; Multi-Scale Fractal Dimension; Fourier Transform.
\\


\section{Introduction}
\label{sec:introduction}
\noindent In pattern recognition and image analysis, shape is one of the most important visual attributes to characterize objects. It provides the most relevant information about an object in order to perform its identification and classification. It is a classical problem, and literature presents a large amount of techniques to extract information related to shape geometric aspect, allowing to separate and to label different parts of an image \citep{Loncaric98, Torres2003}. 

An approach widely used in shape analysis applications is the study of shape through its complexity. In this analysis, complexity is straight related to the irregularity pattern presented by the shape under analysis and, respectively, the amount of the space the shape occupies \citep{journals/pami/ChaudhuriS95,citeulike:591845}.

An interesting way to estimate the complexity of an object is using the Fractal Dimension \citep{Mand2000}. Different of topological dimension, which is an integer number, the Fractal Dimension is a fractionary value that describes how irregular is an object and how much of the space it occupies. Bouligand-Minkowski method is one of the most accurate methods to compute Fractal Dimension. Is is based on the study of the shape influence area computed by shape dilation \citep{books/tricot,books/fm/Schroeder96}.

An alternative method for estimating shape complexity is the Multi-Scale Fractal Dimension. It consists in estimating a curve that represents the changes in complexity shape as we change the visualization scale. Different of Fractal Dimension, which is a numeric value, this approach produces a curve which performs a more accurate shape discrimination \citep{books/tricot,bb21788,GW2002,plotzecanadian}.

In this paper, we propose to evaluate both Fractal Dimension and Multi-Scale Fractal Dimension in a shape analysis context. For this, a shape database is used and experiments are performed considering different configurations of each method. This paper is organized as follows: Section \ref{sec:fractal_dimension} describes the Fractal Dimension method while section \ref{sec:multiscale_fractal_dimension} shows how to compute the Multi-Scale Fractal Dimension. In section \ref{sec:signature} is showed how to use the Multi-Scale Fractal Dimension as a shape signature. The signature making process is based on Fourier analysis and it is described in section \ref{sec:fourier}. Section \ref{sec:experiments} describes the methodology adopted by this work. Results and conclusion are finally given in sections \ref{sec:results} and \ref{sec:conclusion}, respectively.

\section{Fractal Dimension}
\label{sec:fractal_dimension}
\begin{figure}[!ht]
\centering
\includegraphics[scale=0.6]{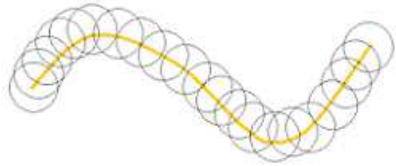}
\caption{Dilation of a shape with a disc of radius $r$.}
\label{fig:dilatacao}
\end{figure}

\noindent Fractal Dimension is a measure of how fragmented a fractal object is, and it may be understood as a characterization of its self-similarity \citep{Mand2000}. It is a non-integer number that quantifies the density of fractals in the metric space and it is a way to identify how complex a fractal is, in order to compare it with another.
Fractal geometry has various approaches to compute the Fractal Dimension of an object. These approaches can be classified as belonging to the Hausdorff-Besicovitch Dimension (like the BoxCounting and Dividers methods) or to the Bouligand-Minkowski Dimension (Minkowski Fractal Dimension method), where the last is the one which produces the most accurate and consistent results for Fractal Dimension \citep{books/tricot,books/fm/Schroeder96,bb16963}. 
Bouligand-Minkowski Fractal Dimension method is based on the study of the influence area created by shape dilation using a disc of radius $r$ (Figure \ref{fig:dilatacao}). Small modifications on the shape produce modifications in the computed influence area \citep{books/tricot,books/fm/Schroeder96}. 

Consider $A \in R^2$ the shape under analysis, the dilation of $A$, $A(r)$, is defined as the set of points in $R^2$ such distance from $A$ is smaller than or equal to $r$:

$$A(r) = \left\{ x \in R^2 | \exists y \in A: \left| x - y \right| \leq r \right\}.$$
 
This dilation can also be defined as:

$$A(r) = \bigcup_{x \in A}B_r(x),$$
 
where $B_r(x)$ is a disc of radius $r$.
The influence area, $A(r)$, and the radius, $r$, follows the relation:
$$A(r) = \mu r^{2-D}.$$ 
So, the Fractal Dimension can be estimated as:  

$$D = 2 - \lim_{r \rightarrow 0} \frac{\log A(r)}{\log r}$$
 
where $D$ is the Fractal Dimension estimated by the Bouligand-Minkowski method.
Through line regression of log-log curve $A(r)$, it is possible to calculate a line with $\alpha$ slope, where $D = 2 - \alpha$ is the Fractal Dimension of the shape using the Bouligand-Minkowski method \citep{books/tricot}.

\begin{figure}[!ht]
\centering
\begin{tabular}{cc}
\includegraphics[scale=0.5]{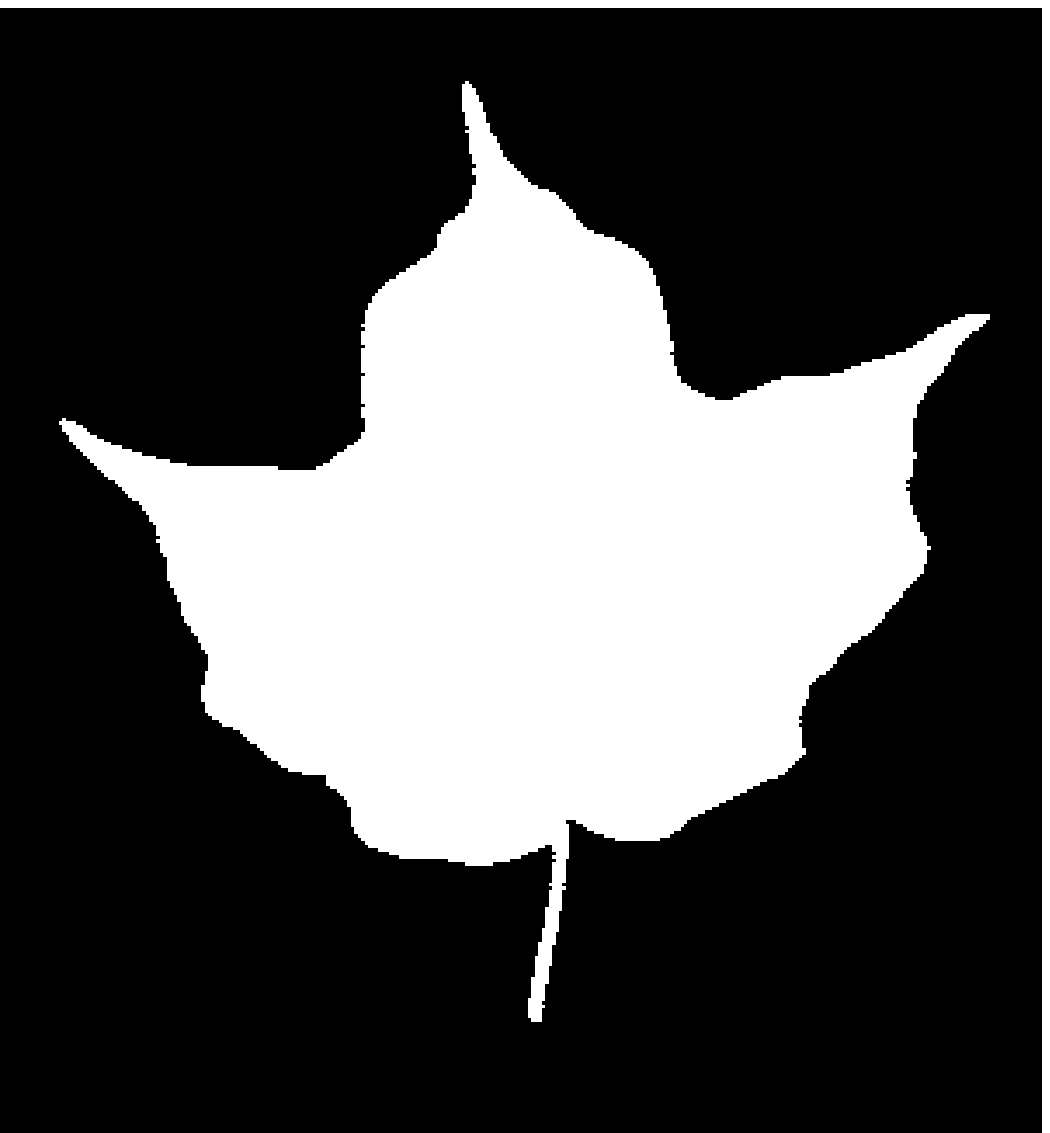} &
\includegraphics[scale=0.5]{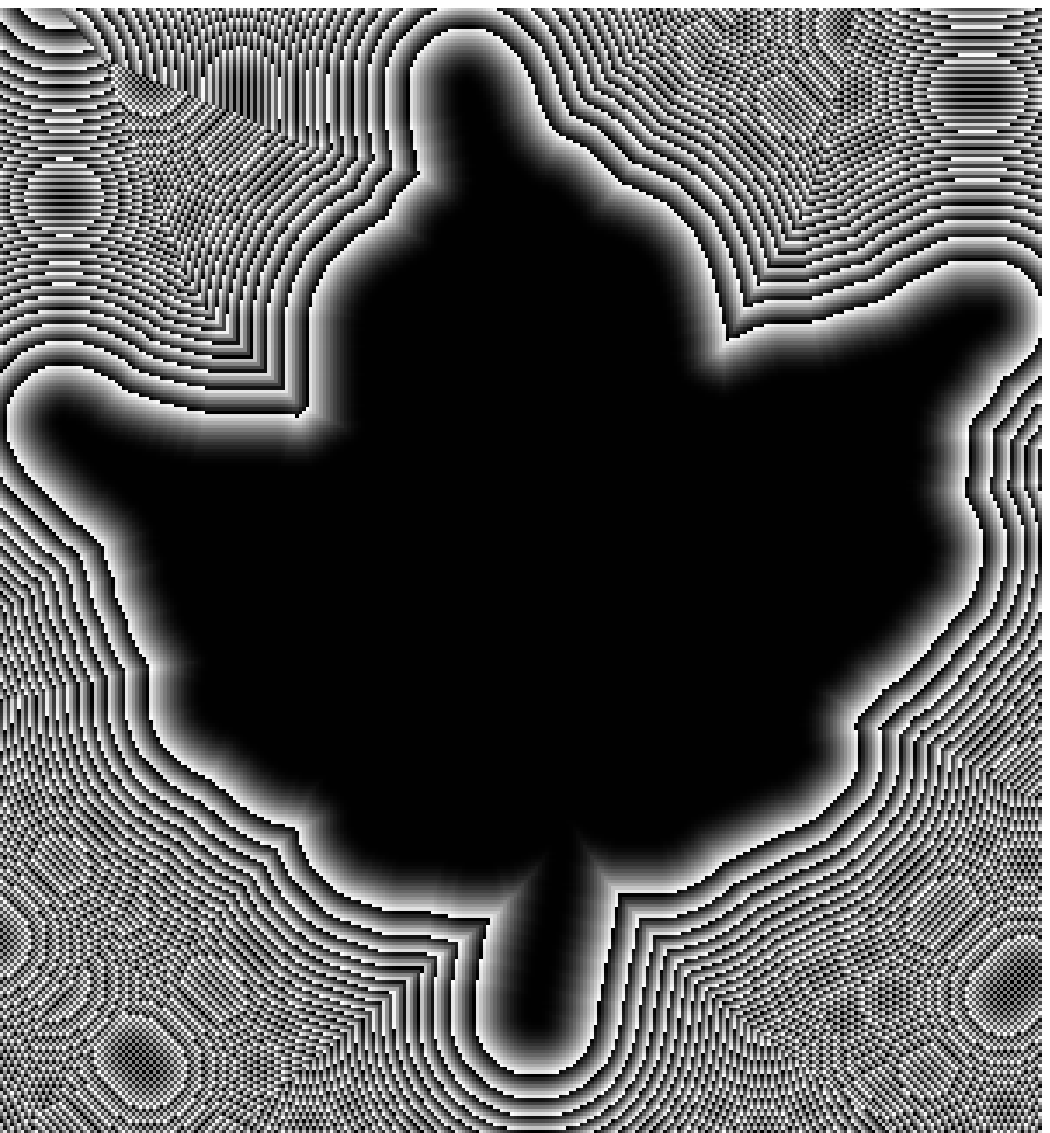}\\
\end{tabular}
\caption{Euclidean Distance Transform ($EDT$) applied over an binary image.}
\label{fig:folha_EDT}
\end{figure}

Computing the influence area of an object is a task of high processing cost. One possibility to optimize this task is using the Euclidean Distance Transform ($EDT$), which attributes to the pixels of a binary image the minimum distance among pixels from the image object to the pixels from the image background. The distance function considered during the $EDT$ computing is the Euclidean distance, due its rotation invariance (Figure \ref{fig:folha_EDT}) \citep{journals/mam/BrunoC04,journals/pr/SaitoT94}.

\section{Multi-Scale Fractal Dimension}
\label{sec:multiscale_fractal_dimension}
\noindent The main problem with Fractal Dimension in nature shape characterization is that they are not fractals (or self-similar). All objects have a finite size, and it implies that their complexities go to zero as long as observation scale increases. An interesting interpretation of the fractal behavior of a shape is how the dilation, using the Bouligand-Minkowski method, occurs in different points of the shape. Some shapes allow their points to be freely dilated while in other points this dilation is saturated in some radius \citep{books/tricot,bb21788}. 

\begin{figure}[!ht]
\centering
\begin{tabular}{cc}
\includegraphics[scale=0.47]{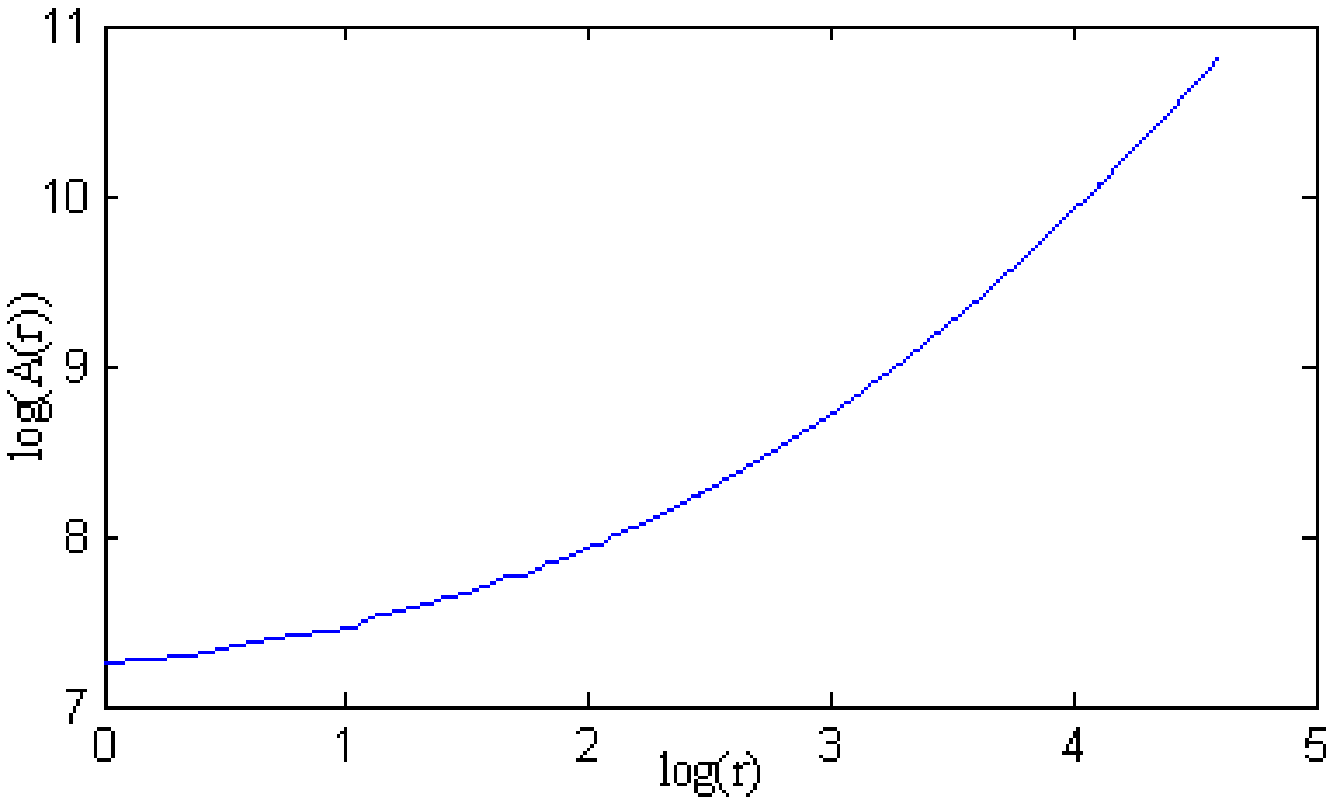} & \includegraphics[scale=0.47]{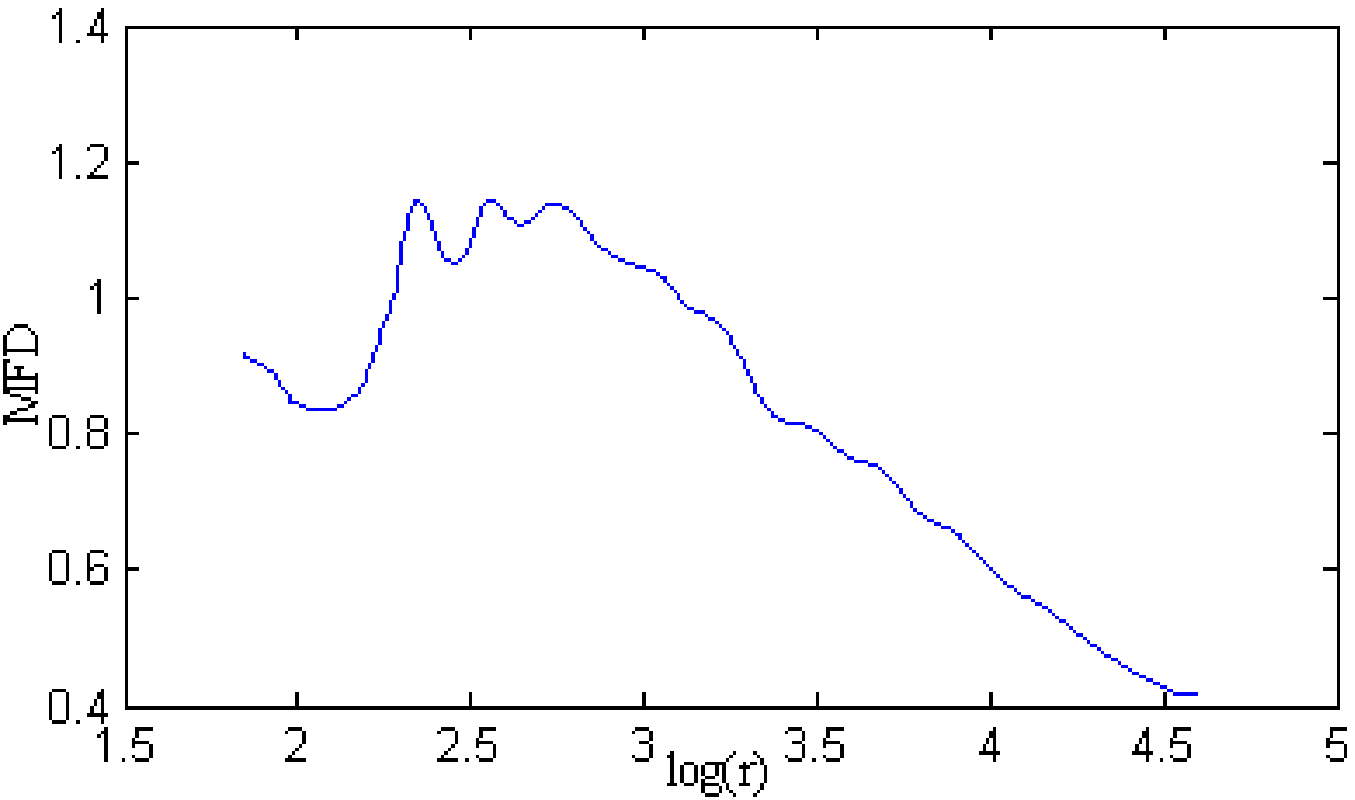}\\
(a) & (b) \\
\end{tabular}
\caption{(a) log-log curve from Bouligand-Minkowski method. (b) Multi-scale Fractal Dimension.}
\label{fig:DFM}
\end{figure}

This behavior gives to the Bouligand-Minkowski log-log curve a richness of details that can not be expressed by just a numeric value, as performed using line regression.
Using the derivate, it is possible to find a function that binds the Fractal Dimension changes to the dilation radius changes (Figure \ref{fig:DFM}) \citep{bb21788,GW2002,plotzecanadian}. This function is called Multi-Scale Fractal Dimension ($MFD$), and it is defined as:	
 
$$MFD = 2 - \frac{du(t)}{dt}$$

where $du(t)/dt$ is the derivative of log-log curve $u(t)$ calculated by Bouligand-Minkowski method. 
 
In order to compute the $MFD$, it is necessary to calculate the derivative of $u(t)$. This is performed using the derivative property of the Fourier Transform. This property allows to compute the derivative curve in the spectrum and it has a better performance when compared with numeric methods, once it considers all data points during the derivative computing \citep{bb21788,plotzecanadian}.
An important detail that requires attention is that derivative methods have a tendency to emphasize high frequency noise. So, it is necessary to use a low pass filter, like Gaussian filter, in order to reduced the this noise \citep{bb16963,Brigham-1988}.
The derivative based on Fourier Transform can be expressed as:
 
$$\frac{du(t)}{dt} = F^{-1}\left\{F\left\{u(t)\right\} F\left\{g_{\sigma}(t)\right\}(j2\pi f) \right\}$$
with
$$g_{\sigma}(t) = \frac{1}{\sigma\sqrt{2\pi}} \exp\left(\frac{-t^2}{2\sigma^2}\right),$$

where $t$ and $u(t)$ are, respectively, the logarithm of radius and the influences area from Bouligand-Minkowski method, $f$ is the frequency, $j$ is the imaginary number and $g_{\sigma}(t)$ is the Gaussian function with standard deviation $\sigma$ \citep{bb21788,GW2002,plotzecanadian}.

\begin{figure}[!htb]
\centering
\includegraphics[scale=0.6]{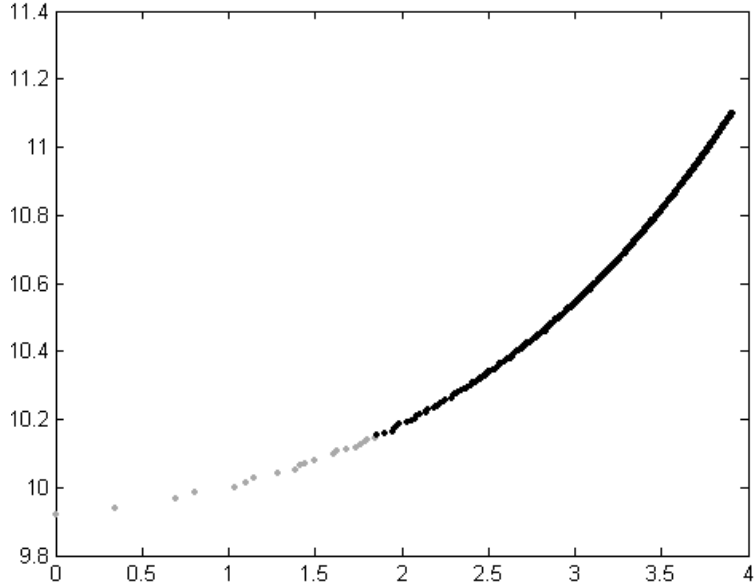}
\caption{In light gray: low sampling region where the points are not considered.}
\label{fig:baixa_amostragem}
\end{figure}

Although the advantages of the derivative property of the Fourier Transform, some important aspects must be considered when computing the $MFD$ curve. These aspects regards to the log-log curve $u(t)$ behavior and they are necessary to achieve a good shape discrimination. At first, it is necessary to provide a curve with a good sampling and uniform interval. This is performed when the initial points of the curve are not considered, once they present low sampling (Figure \ref{fig:baixa_amostragem}), followed by a linear interpolation, which is performed by filling the space between each two points of the sampled curve by its average point.

\begin{figure}[!htb]
\centering
\includegraphics[scale=0.6]{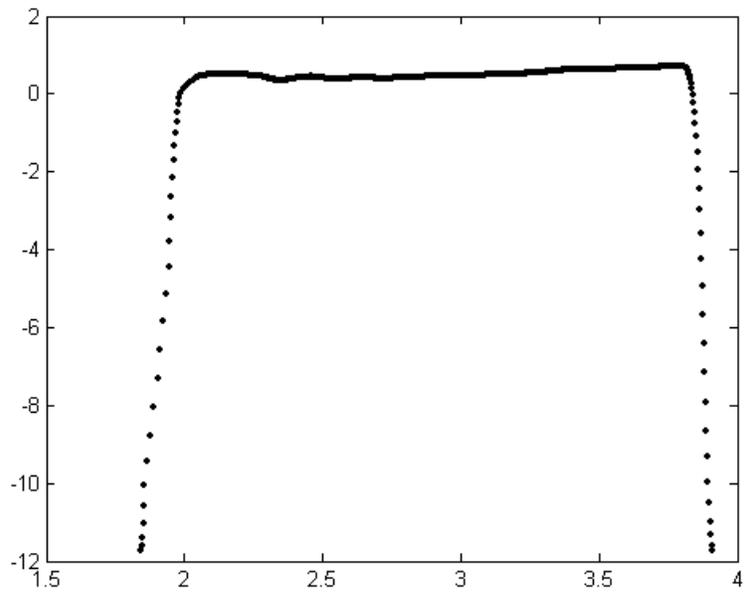}
\caption{Gibbs phenomenon in the curve limits.}
\label{fig:fenomeno_gibbs}
\end{figure}

Another problem present in the derivative property of the Fourier Transform is the discontinuity of the method in the curve limits (Figure \ref{fig:fenomeno_gibbs}). This phenomenon is known in literature as Gibbs phenomenon \citep{Brigham-1988}. This phenomenon is due to the fact of the Fourier transform do not converge uniformly in discontinuities.
 
\begin{figure}[!htb]
\centering
\includegraphics[scale=0.6]{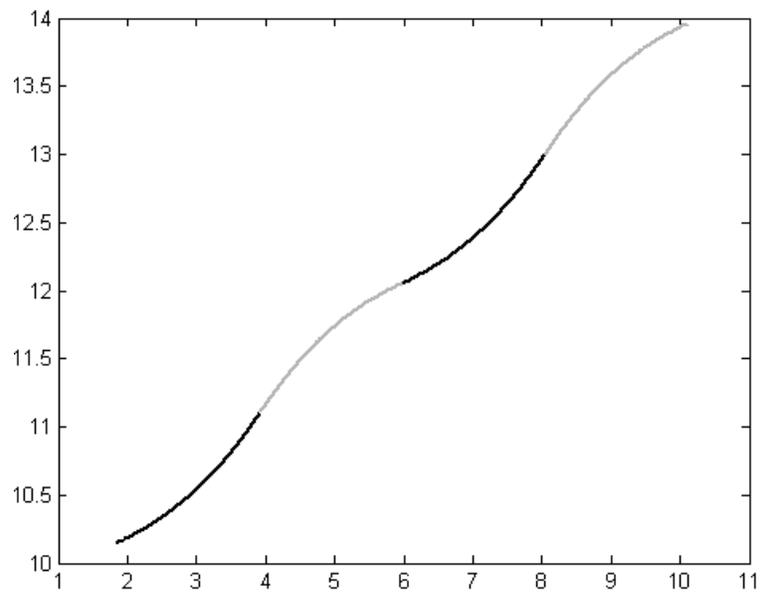}
\caption{Curve duplication and reflexion scheme to provide a continuous curve.}
\label{fig:replicacao}
\end{figure}

An effective solution for this problem is to use a scheme of duplication and reflexion of the curve, so that, it is possible to make it continuous (Figure \ref{fig:replicacao}). This scheme provides a continuous curve for the interval $[2N-1, 3N]$, where $N$ is the length of the original curve.
 
\section{Multi-Scale Fractal Dimension as a shape signature}
\label{sec:signature}
\noindent Image signature is defined as a simplified function or matrix that is able to represent or characterize the original image. In general, techniques used for this task apply a $T:I^2 \rightarrow R$ or $T:I^2 \rightarrow I$ transformation, where the computed feature vector represents the original image in a simplified way. More theory and examples can be found in \citep{GW2002,castl96}.

\begin{figure}[!htb]
\centering
\includegraphics[scale=0.5]{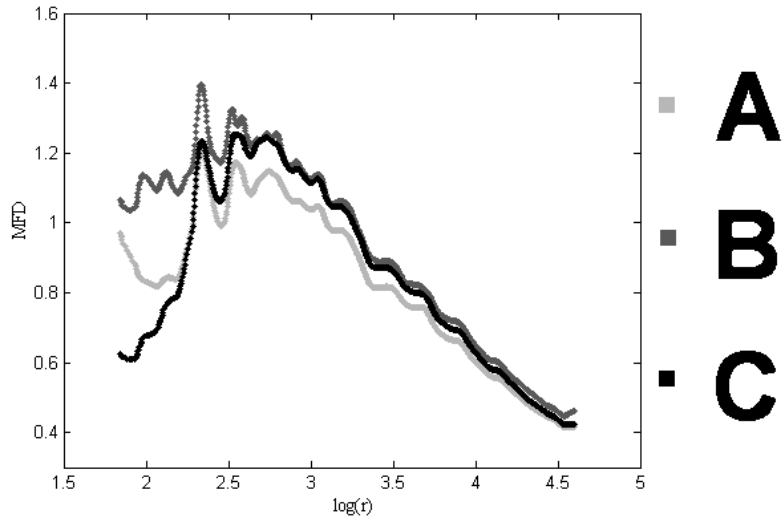}
\caption{$MFD$ curve as a shape signature.}
\label{fig:ABC}
\end{figure}

This paper presents a study about the possibility of using the $MFD$ technique as a shape complexity signature. Different of true fractals, images from nature are not self-similar. They may look different whether we change the visualization scale and, consequently, their Fractal Dimension is also dependent on the used scale. The $MFD$ technique allows to study the shape behavior through scales. As a result, a 1-D function is calculated for a given shape, where this function describes how the Fractal Dimension of that shape changes as we change the scale (Figure \ref{fig:ABC}).

\section{Fourier Descriptors}
\label{sec:fourier}
\noindent The Fourier transform is a useful technique that allows to study the behavior of a signal through its frequency spectrum. It is widely used in pattern recognition tasks, once it presents a great number of advantages, such: noise tolerance, analysis of a signal into different groups of frequencies and easy data normalization, yielding data which is invariant to rotation, translation and scale \citep{Brigham-1988,Bracewel:2000}. 

When it is applied over a signal $u(t)$, the Fourier transform obtains the complex components $U(f)$:

$$ U(f) = \int^{\infty}_{-\infty} u(t)e^{-j2\pi u}dt  ,$$

which are the signal representation in the frequency spectrum. This approach allows to split the original signal into different groups of frequencies, each one with distinct signal features.

Lower frequency coefficients, for example, are associated to the portion of the spectrum that describes the most relevant information about the signal behavior, whereas higher frequency coefficients possess information about noise and details presented in the signal \citep{bb16963}. For the $MFD$, only the lower frequency coefficients are considered. The Fourier descriptors, $DU(f)$, are computed from the magnitude of the selected low frequency coefficients:

$$ DU(f) = \left\| U(f) \right\| .$$

A normalization is also performed in such descriptors, and it is done as follows: 

$$ DU(f) = DU(f) / DU(1).$$

This normalization process increases the tolerance of the descriptors to disturbances in the original signal, such as changes in scale, translation and rotation.

\section{Experiment}
\label{sec:experiments}
\noindent Considering the possibility of using the $MFD$ as a shape signature, an experiment based on image classification is performed to evaluate its efficiency. 

An image database with 1352 images previously classified into 26 classes of 52 images is employed. Each class corresponds to an uppercase letter of the occidental alphabet (26 letters latin alphabet). Images from each class are grouped into 4 groups of 13 images, where each group corresponds to a different random noise level. This allows to evaluate the noise tolerance of the method.

\begin{figure}[!htb]
\centering
\begin{tabular}{cc}
(a) &  \includegraphics[scale=0.7]{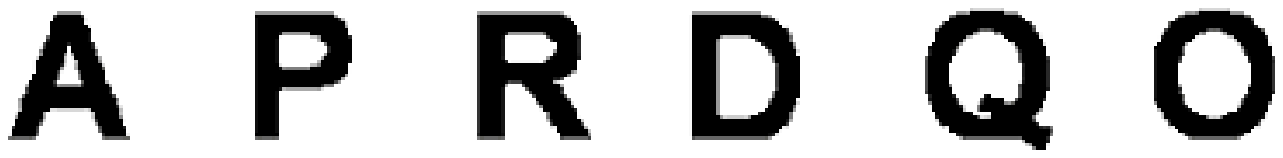}\\
(b) &  \includegraphics[scale=0.7]{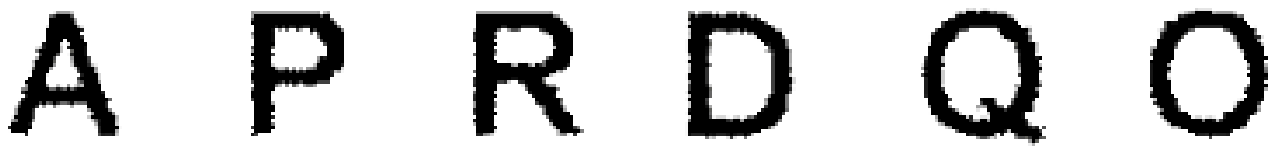}\\
(c) &  \includegraphics[scale=0.7]{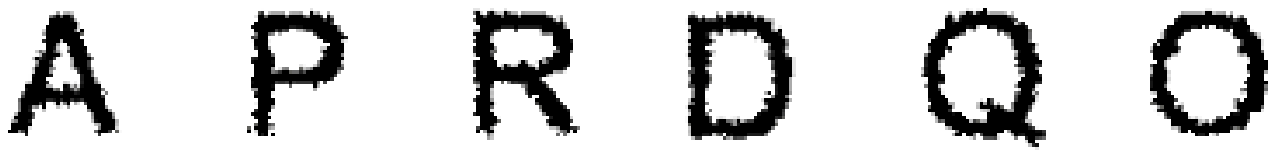}\\
(d) &  \includegraphics[scale=0.7]{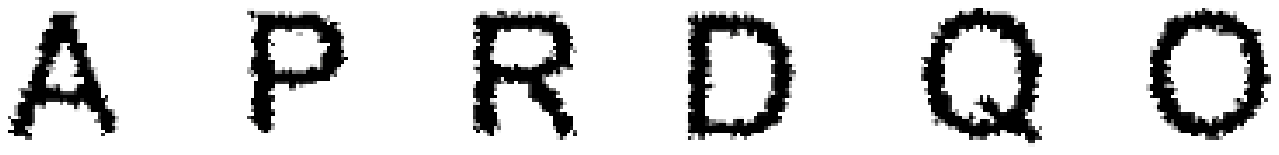}\\
(e) &  \includegraphics[scale=0.7]{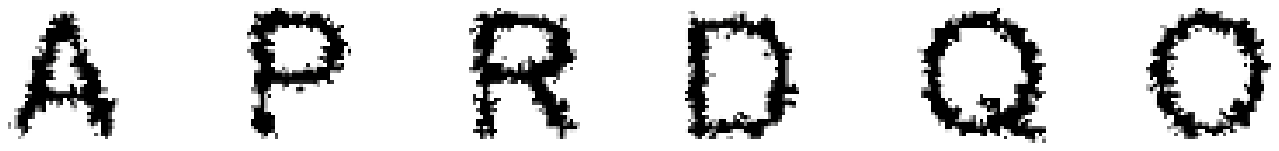}\\
\end{tabular}
\caption{Example of noise levels: (a) Original Shape; (b) Level 1; (c) Level 2; (d) Level 3; (e) Level 4.}
\label{fig:letrasRuido}
\end{figure}

In order to achieve the best performance of the method, various parameters values have been tested. These parameters refers to the dilation radius $r$ used in Bouligand-Minkowski method and the smoothing level $\sigma$ used in the $MFD$ calculation. For each different configuration, a new $MFD$ curve is computed for each sample in the image database. An average $MFD$ curve is also computed for each class using 12 samples of it (3 samples for each noise level), and these used curves are discarded in order to not affect the classification results. So, the test database is compound of 1040 image, divided into 26 classes of 40 images. Each class is also divided into 4 noise levels, each one with 10 images. Classification process is performed by Euclidean distance, where it is verified the distance among the remainders curves and the average $MFD$. 

\section{Results and Discussion}
\label{sec:results}
\noindent In this section some properties of $MFD$ curve are discussed, as well as its performance according to its parameter. Besides, we also compare the results from $MFD$ and Fractal Dimension, and the advantages and disadvantages of using Fourier Descriptors from $MFD$ curve for shape classification.

\subsection{Method parameters analysis}
\noindent In order to attain the best method performance, various parameters values have been tested. The tested parameters values are $\sigma \in \left\{10,15,20,25 \right\}$ and \linebreak
$r \in \left\{10,15,20,25,30,35,40,45,50,75,100,125,150,175,200,225 \right\}$. The results for each different configuration showed the characteristics and properties present in $MFD$ curve.

\begin{figure}[!htb]
\centering
\includegraphics[scale=0.5]{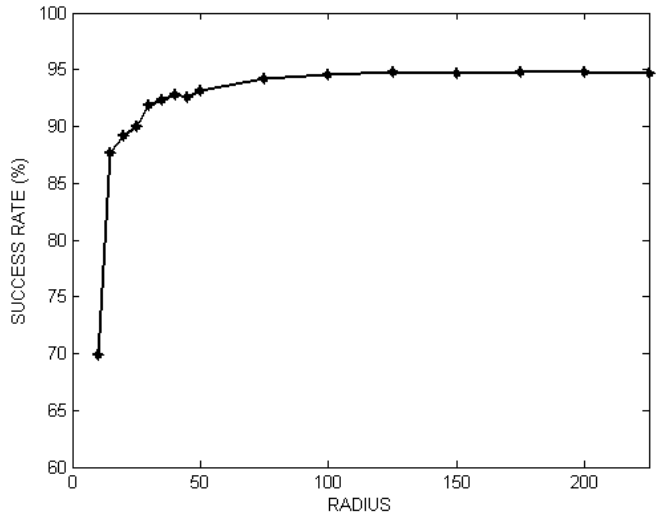}
\caption{$MFD$ success rate as radius increases.}
\label{fig:AcertosMultiescala}
\end{figure}

\begin{figure}[!htb]
\centering
\begin{tabular}{cc}
\includegraphics[scale=0.75]{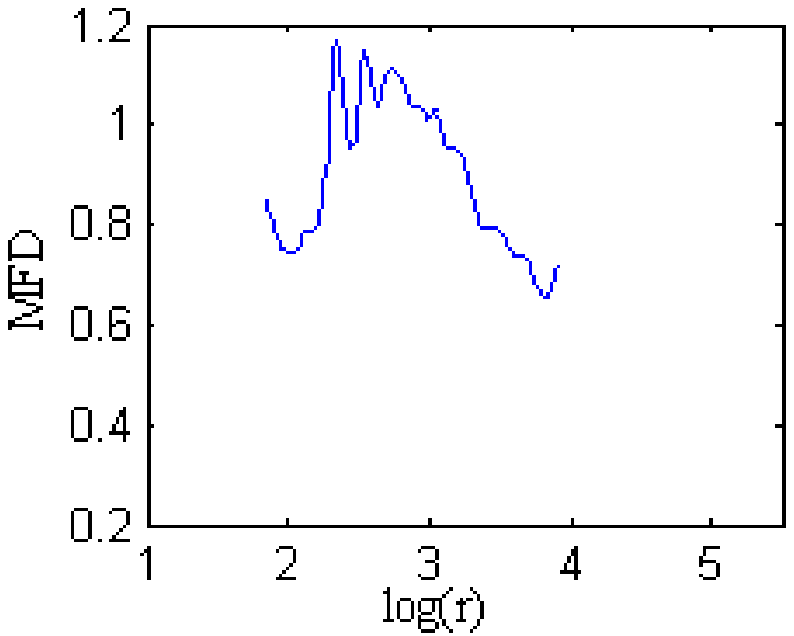} & \includegraphics[scale=0.75]{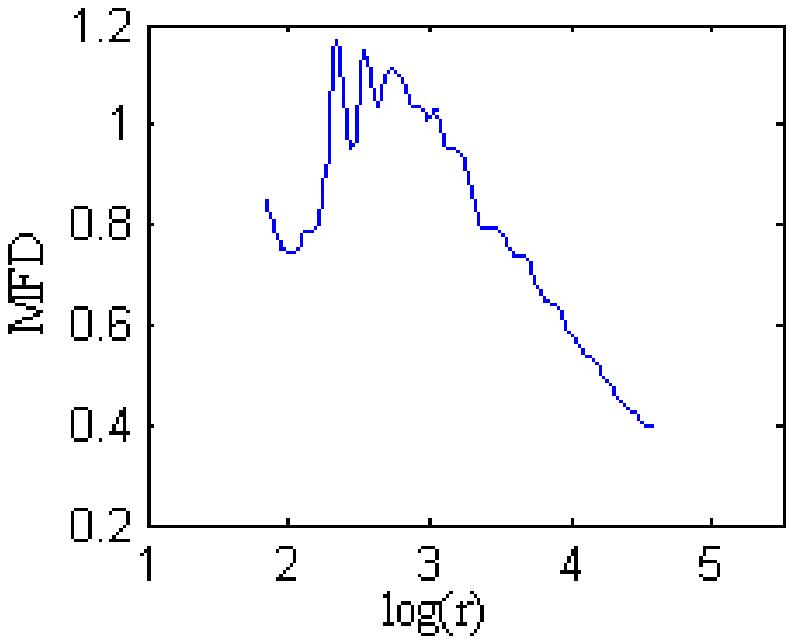}\\
(a) & (b)\\
\includegraphics[scale=0.75]{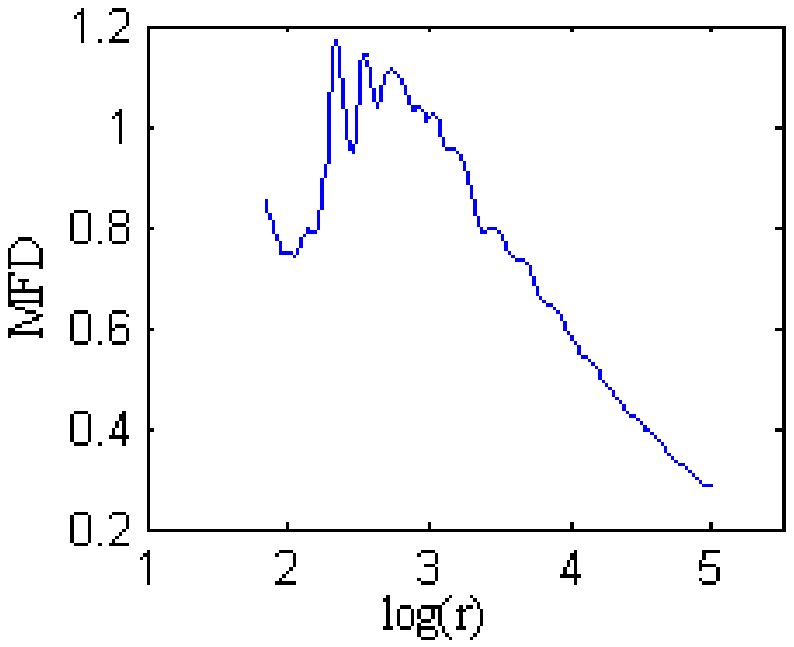} & \includegraphics[scale=0.75]{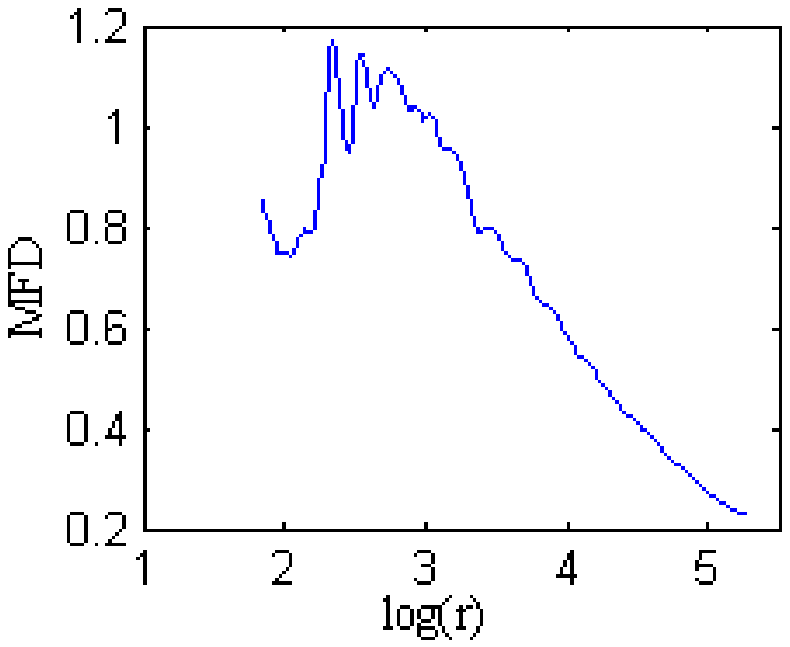}\\
(c) & (d)\\
\end{tabular}
\caption{Example of $MFD$ computed for different dilation radius $r$, using $\sigma = 10$: (a) $r = 75$; (b) $r = 100$; (c) $r = 125$; (d) $r = 150$.}
\label{fig:RaiosDilatacao}
\end{figure}

We note an increase in success rate as $r$ increases (Figure \ref{fig:AcertosMultiescala}). However, the relation between success rate and the dilation radius presents a log-curve behavior. So, from an specific radius value, the success rate becomes constant. The explanation for this behavior lies in the fact that as $r$ increases, the dilation produced by Bouligand-Minkowski method makes the shape aspect becomes more similar to a point, whose Fractal Dimension is equal to zero. This behavior is represent in $MFD$ curve as its tendency to zero according to $r$ increase. From an specific $r$, none relevant information about the shape is added to $MFD$ curve, except for its zero-tendency, what keeps the success rate stable. We also note that this behavior do not depends on $\sigma$ value (Figure \ref{fig:RaiosDilatacao}).

\begin{figure}[!htb]
\centering
\begin{tabular}{cc}
\includegraphics[scale=0.65]{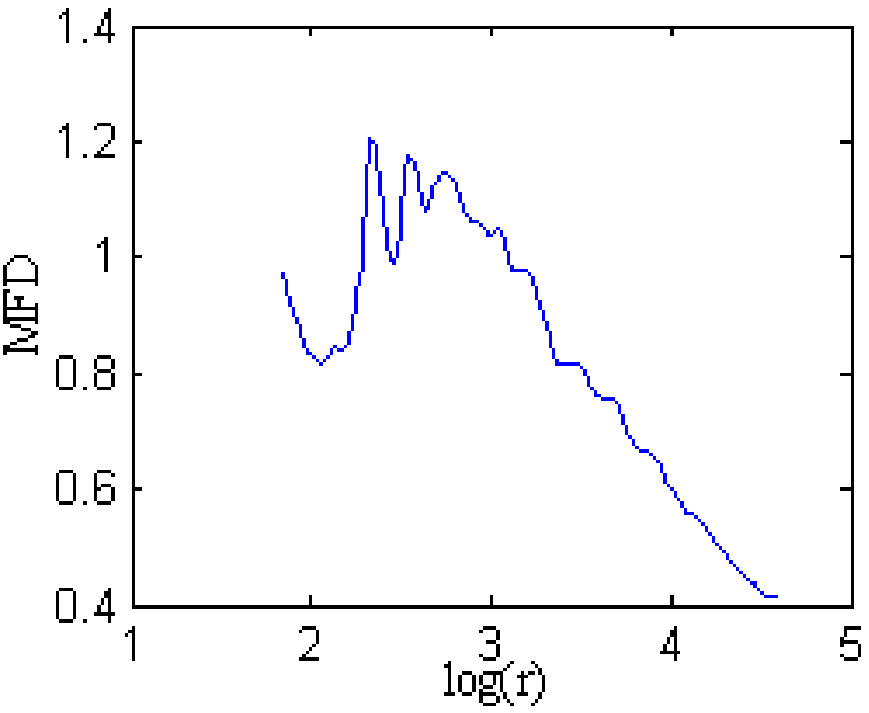} & \includegraphics[scale=0.65]{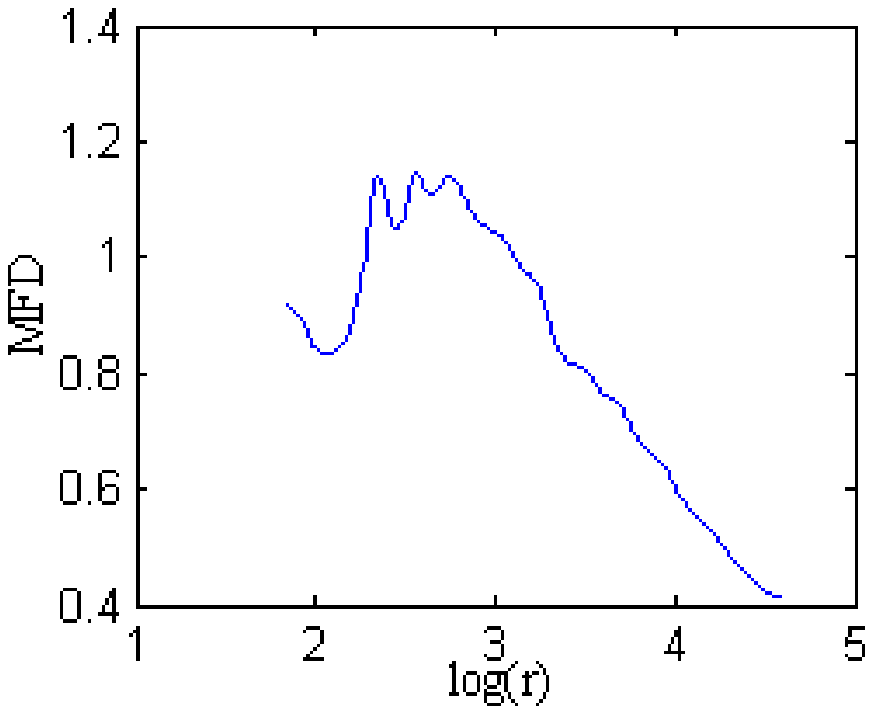}\\
(a) & (b)\\
\includegraphics[scale=0.65]{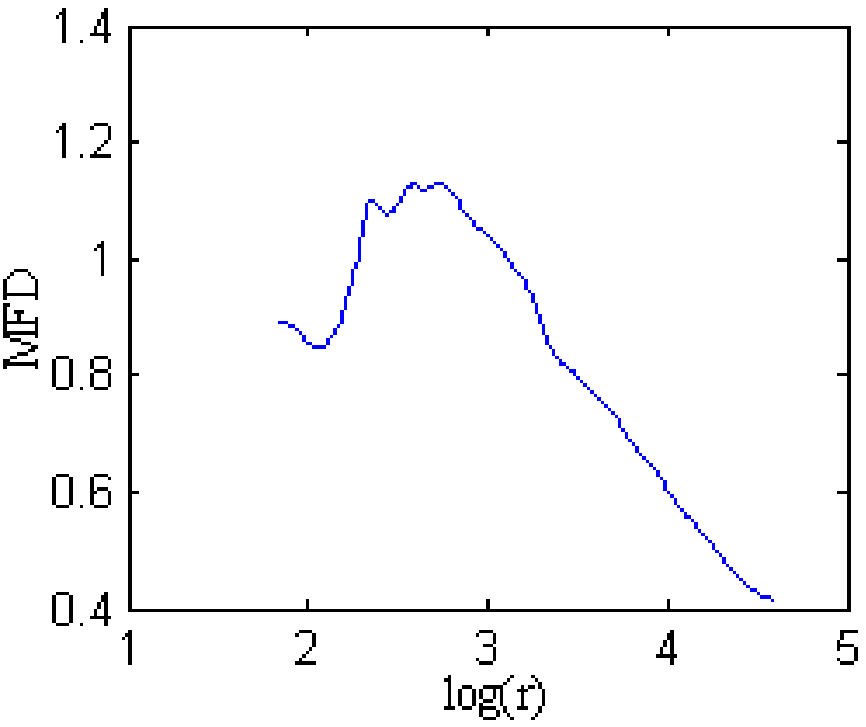} & \includegraphics[scale=0.65]{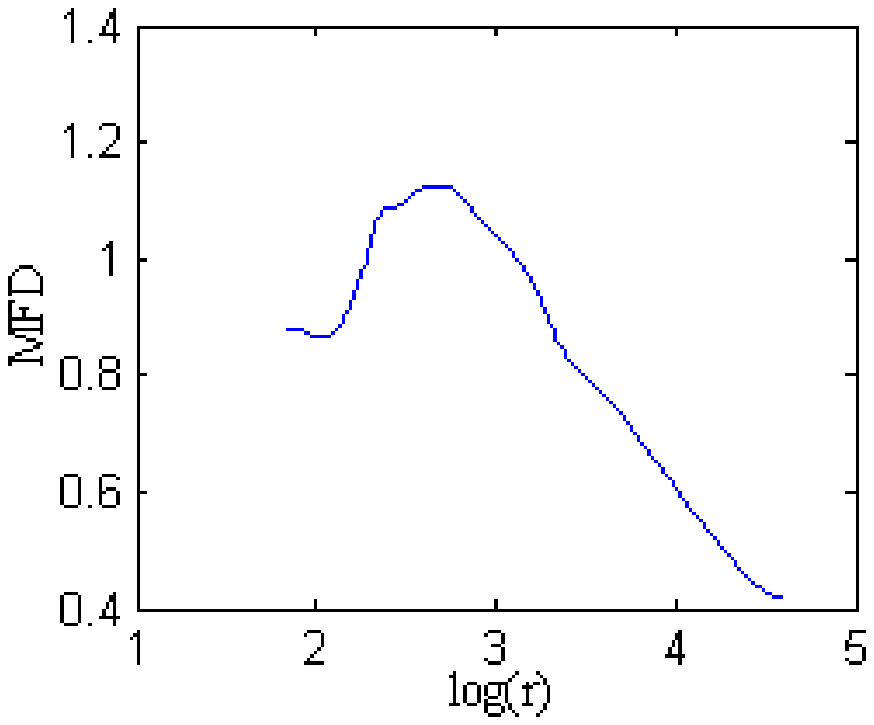}\\
(c) & (d)\\
\end{tabular}
\caption{Example of $MFD$ computed for different $\sigma$ values, using $r = 100$: (a) $\sigma = 10$; (b) $\sigma = 15$; (c) $\sigma = 20$; (d) $\sigma = 25$.}
\label{fig:VariaSigma}
\end{figure}

\begin{figure}[!htb]
\centering
\includegraphics[scale=0.5]{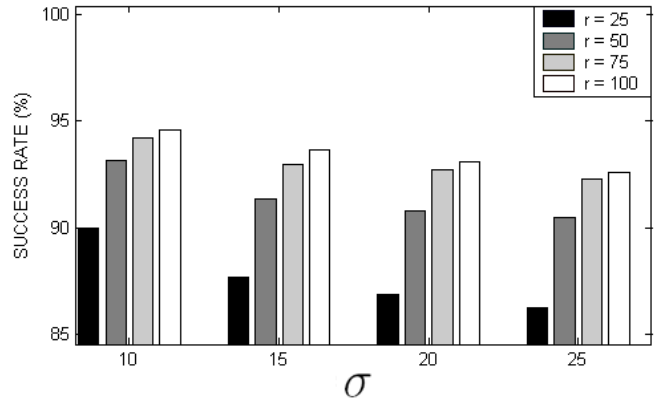}
\caption{$MFD$ success rate for different $\sigma$ values.}
\label{fig:AcertosSigma}
\end{figure}

Besides $r$ parameter, which refers to dilation radius, we realize that $\sigma$ value also carry out some influence over the success rate. The parameter $\sigma$ refers to the smoothing level of $MFD$ curve and it is responsible for the presence of more or less details in the curve (Figure \ref{fig:VariaSigma}). We note higher values for this parameter produce a excessive smoothing and it suppresses important details of the curve. As a consequence, the success rate decreases, independent of the value of parameter $r$ (Figure \ref{fig:AcertosSigma}).

In this experiment, the best performance is achieved when used $r = 100$ and $\sigma =10$.

\begin{table}[!htb]
\setlength{\tabcolsep}{3.0pt}
\renewcommand{\arraystretch}{1}
\centering
\tiny
\begin{tabular}{c|cccccccccccccccccccccccccc}
&A&B&C&D&E&F&G&H&I&J&K&L&M&N&O&P&Q&R&S&T&U&V&W&X&Y&Z\\
\hline
A&38&&&&&2&&&&&&&&&&&&&&&&&&&&\\
B&&40&&&&&&&&&&&&&&&&&&&&&&&&\\
C&&&40&&&&&&&&&&&&&&&&&&&&&&&\\
D&&&&40&&&&&&&&&&&&&&&&&&&&&&\\
E&&&&&38&&&&&&&&&&&&&1&1&&&&&&&\\
F&&&&&&39&&&&&&&&&&&&&&&&1&&&&\\
G&&&&&&&40&&&&&&&&&&&&&&&&&&&\\
H&&&&&&&&40&&&&&&&&&&&&&&&&&&\\
I&&&&&&&&&40&&&&&&&&&&&&&&&&&\\
J&&&&&&&&&&36&&1&&&&&&&&1&&&&&2&\\
K&&&&&&&&&&&36&&&&&&&&&&&&&&&4\\
L&&&&&&&&&&1&&35&&&&&&&&4&&&&&&\\
M&&&&&&&&&&&&&40&&&&&&&&&&&&&\\
N&&&&&&&&&&&&&&30&&&&10&&&&&&&&\\
O&&&&&&&&&&&&&&&40&&&&&&&&&&&\\
P&&&&&&&&&&&&&&&&40&&&&&&&&&&\\
Q&&&&&&&&&&&&&&&&&40&&&&&&&&&\\
R&&&&&&&&&&&&&&6&&&&34&&&&&&&&\\
S&&&&&&&&&&&&&&&&&&&40&&&&&&&\\
T&&&&&&&&&&&&11&&&&&&&&27&&&&&2&\\
U&&&&&&&&1&&&&&&&&&&&&&39&&&&&\\
V&&&&&&&&&&&&&&&&&&&&&&40&&&&\\
W&&&&&&&&&&&&&&&&&&&&&&&40&&&\\
X&&&&&&&&&&&&&&&&&&&&&&&&38&&2\\
Y&&&&&&&&&&1&&&&&&&&&&&&&&&39&\\
Z&&&&&&&&&&&4&&&&&&&&&&&&&1&&35\\
\end{tabular}
\caption{Confusion Matrix for Multi-Scale Fractal Dimension method, considering $r = 100$ and $\sigma = 10$. Success rate = 94.62 \%.}
\label{tab:confusao_dfm}
\end{table}

\begin{table}[!htb]
\setlength{\tabcolsep}{3.0pt}
\renewcommand{\arraystretch}{1}
\centering
\tiny
\begin{tabular}{c|cccccccccccccccccccccccccc}
&A&B&C&D&E&F&G&H&I&J&K&L&M&N&O&P&Q&R&S&T&U&V&W&X&Y&Z\\
\hline
A&11&&&&&&&&&&&&&&&7&&&&&&19&&3&&\\
B&&14&&&&&16&&&&&&&4&&&6&&&&&&&&&\\
C&&&24&&&&&&&&&&&&&&&&6&&&&&6&&4\\
D&&1&&11&3&&&18&&&&&&3&2&&&2&&&&&&&&\\
E&&&&1&7&&&5&&&4&&&&&&&1&10&&12&&&&&\\
F&&&&&&39&&&&&&&&&&&&&&&&1&&&&\\
G&&8&&1&&&9&&&&&&&18&2&&2&&&&&&&&&\\
H&&&&4&13&&&10&&&6&&&1&2&&&2&&&2&&&&&\\
I&&&&&&&&&40&&&&&&&&&&&&&&&&&\\
J&&&&&&&&&&30&&5&&&&&&&&5&&&&&&\\
K&&&&3&3&&&3&&&3&&&&1&&&&20&&6&&&&&1\\
L&&&&&&&&&&19&&9&&&&&&&&9&&&&&3&\\
M&&&&&&&&&&&&&36&&&&&&&&&&4&&&\\
N&&4&&4&&&7&1&&&&&&8&10&&&6&&&&&&&&\\
O&&2&&22&&&2&2&&&&&&4&4&&&4&&&&&&&&\\
P&13&&&&&&&&&&&&&&&14&&&&&&12&&1&&\\
Q&&3&&&&&&&&&&&&&&&37&&&&&&&&&\\
R&&1&&8&1&&1&17&&&&&&5&5&&&2&&&&&&&&\\
S&&&&&2&&&&&&2&&&&&&&&17&&3&&&&&16\\
T&&&&&&&&&&5&&8&&&&&&&&19&&&&&8&\\
U&&&&&4&&&2&&&6&&&&&&&&20&&8&&&&&\\
V&6&&&&&2&&&&&&&&&&2&&&&&&30&&&&\\
W&&&&&&&&&&&&&2&&&&&&&&&&38&&&\\
X&&&6&&&&&&&&&&&&&1&&&&&&&&33&&\\
Y&&&&&&1&&&&&&&&&&&&&&9&&&&&30&\\
Z&&&23&&&&&&&&&&&&&&&&3&&1&&&3&&10\\
\end{tabular}
\caption{Confusion Matrix for Fractal Dimension method, considering $r = 100$. Success rate = 47.40 \%.}
\label{tab:confusao_df}
\end{table}

\subsection{Fractal Dimension versus Multi-Scale Fractal Dimension}
\noindent As previously discussed, the main problem with Fractal Dimension in shape characterization is that it does not work properly with real world restrictions. As we increase the visualization scale, the Fractal Dimension of the object goes to zero. In the other hand, $MFD$ curve emphasizes the details present in log-log curve computed from Bouligand-Minkowski method. It allows to produce a more detailed representation of how complexity changes according to the scale. This representation is much more consistent for analysis and less dependent on noise interferences than a simple numeric value as calculated by $u(t)$ line regression.

\begin{figure}[!htb]
\centering
\includegraphics[scale=0.5]{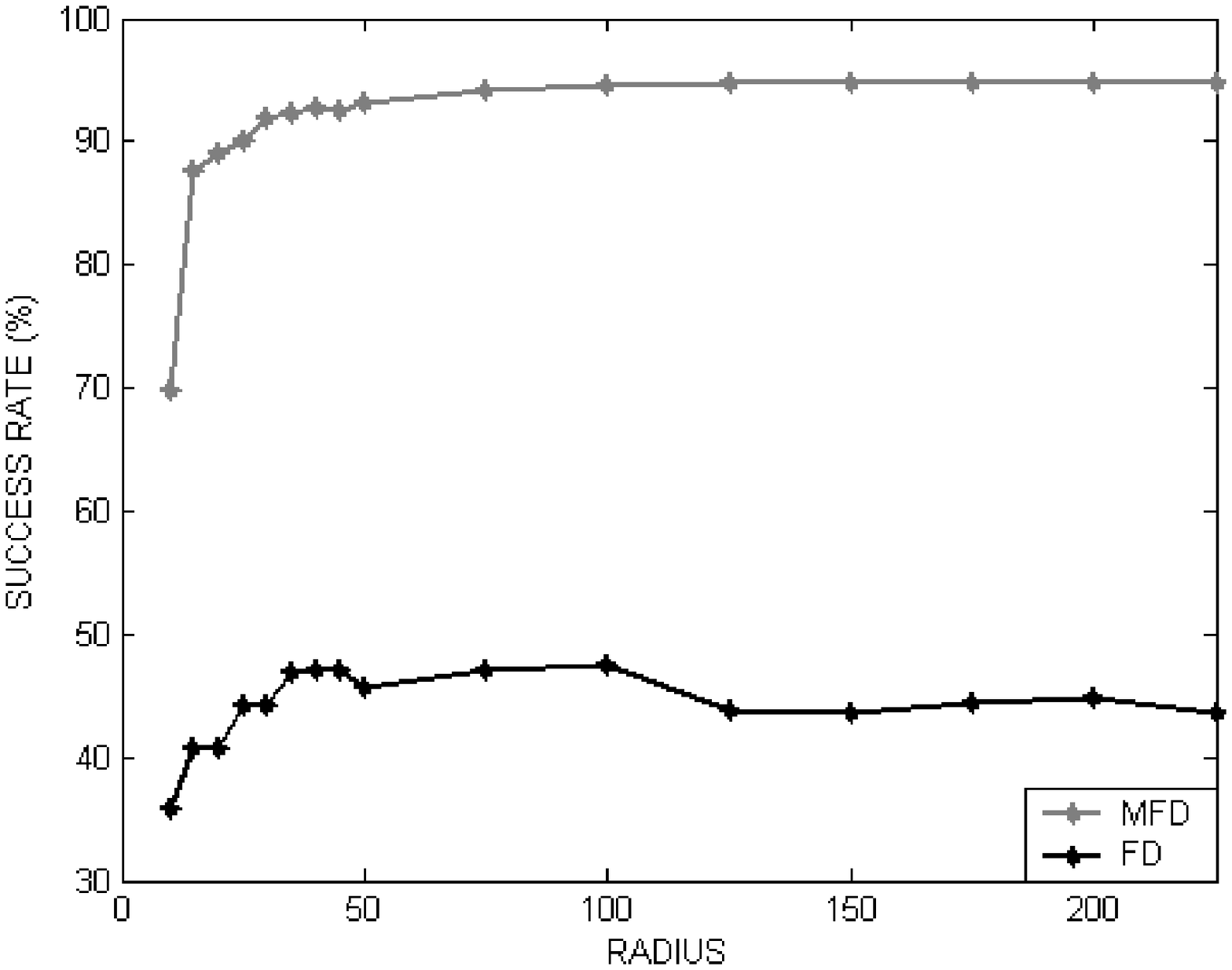}
\caption{$MFD$ and $FD$ success rates as dilation radius increases.}
\label{fig:MFDversusFD}
\end{figure} 

Figure \ref{fig:MFDversusFD} shows the classification results for both methods, Fractal Dimension ($FD$) and Multi-Scale Fractal Dimension ($MFD$). For this experiment, various radius values have been tested and, in $MFD$ case, an $\sigma = 10$ was considered. 

We note a higher success rate for $MFD$ when compared with $FD$, independent of radius values used. In $FD$ method, only an numeric value is calculated from $u(t)$ and used for shape characterization. This becomes the method more sensitive to small variations or noise in the shape. The results from $DF$ are also more unstable than $MFD$. The explanation for this lies in the zero-tendency presented in the Bouligand-Minkowski method. As the dilation radius increases the Fractal Dimension of any object goes to zero, what becomes the classification process more difficult and inaccurate. 

Tables \ref{tab:confusao_dfm} and \ref{tab:confusao_df} show, respectively, the confusions matrix for $DFM$ and $DF$ when considered all images in the test database and $r = 100$.

\subsection{Noise Tolerance}
\noindent An important characteristic to be evaluated in any method is its capacity to work properly with samples that present sort of distortion or noise. In general, images under analysis may present distortions from measuring errors (measures from physical world), or noise added to original data (e.g., interference added during data transmission), that difficults the analysis and classification processes. 

\begin{figure}[!htb]
\centering
\includegraphics[scale=0.5]{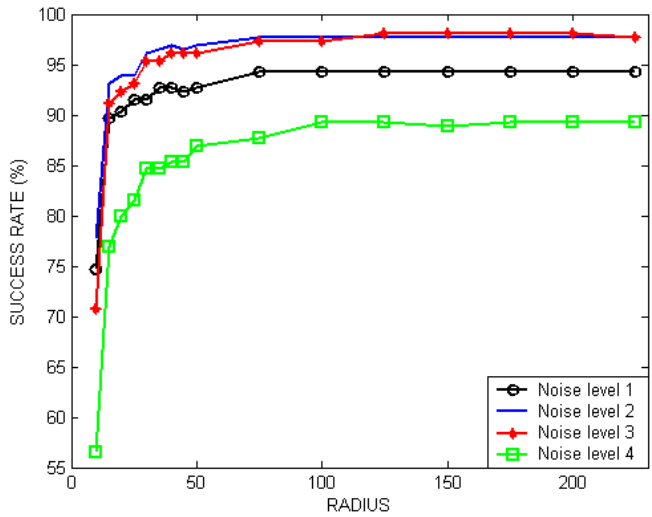}
\caption{$MFD$ success rate at different noise levels and dilation radius.}
\label{fig:MFDnoise}
\end{figure}

By analysis of the results from $MFD$ curve, we note this method present a good noise tolerance. This tolerance comes from using a Gaussian low filter during the computing of $MFD$ curve. The Gaussian filter acts reducing the importance of the informations in the high frequency region of the spectrum. Once noise is a typical high frequency information, it is removed from the original data through this data filtering process. We also note this filtering process presents a better performance on images that present a intermediate noise level, i.e., levels 2 and 3 of the used noise (Figures \ref{fig:letrasRuido}c and \ref{fig:letrasRuido}d, respectively) . In other hand, we also realize a stressed decrease in the performance of the method as the noise exceed these noise levels. High noise levels change the geometric patterns in the shape, what difficults the classification process and, as a consequence, decreases the success rate of the method (Figure \ref{fig:MFDnoise}).

\subsection{Analysis using Fourier descriptors}
\begin{figure}[!htb]
\centering
\includegraphics[scale=0.5]{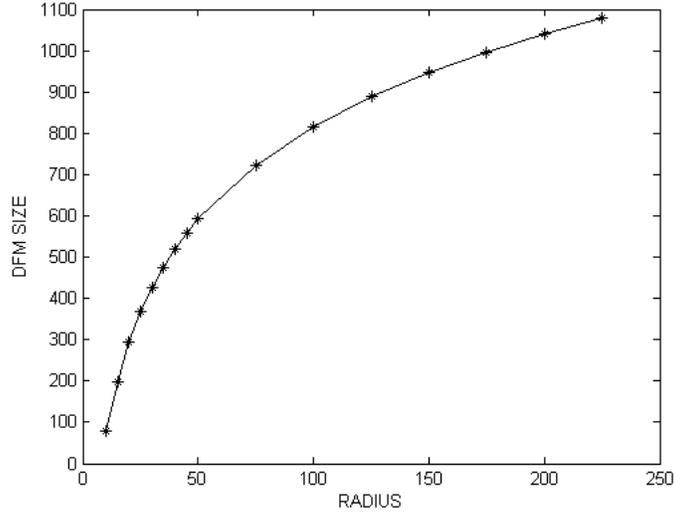}
\caption{Number of descriptors in the $MFD$ curve for the radius used.}
\label{fig:DFM_nro_descriptors}
\end{figure}

\noindent As previously discussed, the main advantage of $MFD$ curve in pattern recognition is its capacity to represent an object by a curve that binds its complexity along the visualization scale. Nevertheless, the $MFD$ curve presents a high cost to be evaluated once its number of descriptors increase as the radius $r$ increases (Figure \ref{fig:DFM_nro_descriptors}). Besides that, not all information in the curve is relevant for shape classification. 

In order to simplify the analysis process the Fourier transform is applied over $MFD$ curve, so descriptors that represent the main behavior of the curve are computed, as mentioned in Section \ref{sec:fourier}. 

\begin{figure}[!htb]
\centering
\includegraphics[scale=0.5]{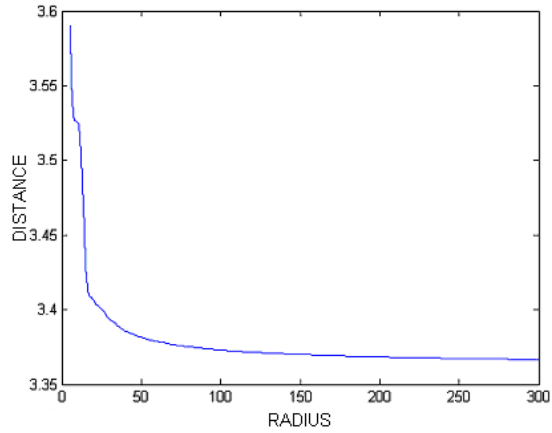}
\caption{Distance between $MFD$ and its Fourier descriptors as the number of computed descriptors increase.}
\label{fig:FourierError}
\end{figure}

One important point to be considered is the number of descriptors to be computed. The number of descriptors have been defined through analysis of the Euclidean distance between the original $MFD$ curve and its respective set of descriptors. Figure \ref{fig:FourierError} shows the variation of this distance as we increase the number of Fourier descriptors. We realize as we increase the number of descriptors smaller becomes the distance between the $MFD$ curve and its descriptors. However, this distance becomes stable when the number of descriptors is higher than 50. This shows that most of the main information lies in the 50 first descriptors. 

\begin{figure}[!htb]
\centering
\includegraphics[scale=0.5]{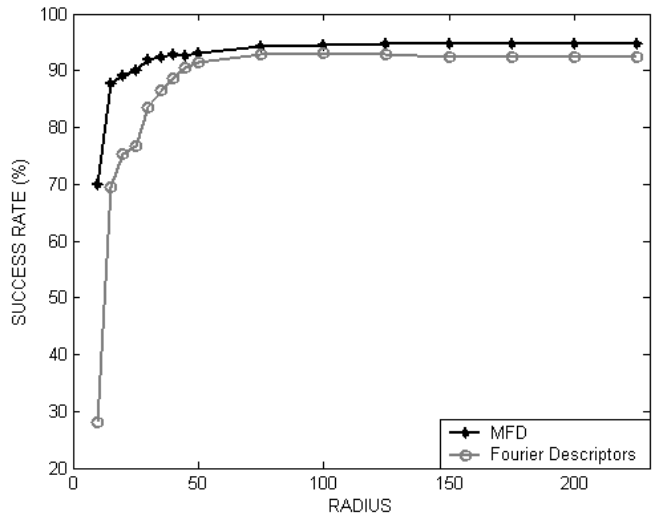}
\caption{$MFD$ and its Fourier descriptors success rate.}
\label{fig:MFDFourier}
\end{figure}

Once it was defined the number of descriptors that best represent the curve behavior, the performance of these descriptors was compared with the original curve in the shape classification experiment. For this, we considered the $MFD$ curve calculated for different dilation radius, $r$, and $\sigma = 10$. Figure \ref{fig:MFDFourier} shows the success rate for both Fourier descriptors and $MFD$ curve.

Using Fourier descriptors is a way to simplify the analysis step, once it reduces the amount of information to be analyzed. Otherwise, suppressing part of curve information, specially high frequency information, produces a smoothing effect over the $MFD$ curve, which decreases the distance between classes and, consequently, decrease the success rate. In spite of the large amount of information discarded during the process of Fourier descriptors computing, the decrease in success rate is minimum for $r \geq 50$. This range of $r$ values is where the method has presented its best results, and it becomes evident the great capacity of $MFD$ curve, in association with Fourier descriptors, to shape classification.

\section{Conclusion}
\label{sec:conclusion}
\noindent 

This paper have presented an experimental comparison between Fractal Dimension and Multi-Scale Fractal Dimension in a shape analysis context. Fourier descriptors have also been performed over Multi-Scale Fractal Dimension curve in order to simplify the analysis step. A previously classified shape database was employed and different configurations of each method have been tested.

Results have shown that Multi-Scale Fractal Dimension performs a more accurate discrimination of the shapes. It also presents a "good" noise tolerance, what is corroborated by experimental results. Experiments also have shown that Fourier descriptors computed from Multi-Scale curve hold the main information of the curve, with minimal loss of information, while it reduces drastically the number of descriptors necessary to shape classification.


\end{document}